%% file: main.tex
\documentclass{article} 
\usepackage{iclr,times}

\input{math_commands.tex}

\usepackage{acronym}
\usepackage{booktabs}
\usepackage{amssymb}

\usepackage{graphicx}
\usepackage{caption}
\usepackage{subcaption}
\usepackage{hyperref}
\usepackage{xcolor}
\usepackage{url}
\usepackage{arydshln} 
\usepackage{amssymb} 
\usepackage{booktabs}
\usepackage{float}
\usepackage{footnote}
\usepackage{multirow}
\usepackage{tikz}
\usepackage{color}
\usepackage{pgfplots}
\usepackage{pgf-umlsd}
\usepackage{ifthen}
\usepackage{forest}

\title{HMCNAS: Neural Architecture Search using Hidden Markov Chains and Bayesian Optimization}

\author{Vasco Lopes, Luís A. Alexandre \\
NOVA LINCS \\
Universidade da Beira Interior, Portugal\\
 \texttt{\{vasco.lopes,luis.alexandre\}@ubi.pt} 
}

\iclrfinalcopy
\begin{document}

\maketitle

\begin{abstract}
Neural Architecture Search has achieved state-of-the-art performance in a variety of tasks, out-performing human-designed networks. However, many assumptions, that require human definition, related with the problems being solved or the models generated are still needed: final model architectures, number of layers to be sampled, forced operations, small search spaces, which ultimately contributes to having models with higher performances at the cost of inducing bias into the system. In this paper, we propose HMCNAS, which is composed of two novel components: i) a method that leverages information about human-designed models to autonomously generate a complex search space, and ii) an Evolutionary Algorithm with Bayesian Optimization that is capable of generating competitive CNNs from scratch, without relying on human-defined parameters or small search spaces. 
The experimental results show that the proposed approach results in competitive architectures obtained in a very short time. HMCNAS provides a step towards generalizing NAS, by providing a way to create competitive models, without requiring any human knowledge about the specific task.
\end{abstract}




\section{Introduction}

The ever-growing success of Machine Learning (ML) applications created a demand for ML systems that can be used off-the-shelf without care for its inner components. This success is mostly attributed to Deep Learning algorithms, especially Neural Networks, which obtained state-of-the-art results in a variety of problems and have been used extensively with great success \citep{lecun2015deep,schmidhuber2015deep}. Deep learning architectures removed the need for extensive hand-crafted feature extraction and pre-processing steps \citep{goodfellow2016deep}. However, applying a ML algorithm to a problem without being explicitly tailor-made for that specific problem, usually results in non-optimal performances, wherein some cases, the models will even yield poor performances. Moreover, designing tailor-made ML algorithms and high performant models can be a difficult task, as there are many design choices that are not independent of each another. This is especially true when talking about Deep Neural Networks, which have parameters associated with the training procedure, the architecture, optimization rules, layers combination and even how to mitigate problems such as overfitting and vanishing or exploding gradient \citep{pascanu2012understanding, srivastava2014dropout}, requiring years of expertize and increasingly demand for architecture engineering. Thus, the need for automating the design of Neural Networks become logical \citep{hutter2019automated}.

Neural Architecture Search (NAS) is a subset of AutoML, that intends to automate architecture engineering \citep{elsken2019neural}. NAS methods have been successfully applied to image classification tasks \citep{cai2018proxylessnas,liu2018darts,zela2020understanding}, semantic segmentation \citep{liu2019auto}, object detection \citep{chen2019detnas}, image generation \citep{gong2019autogan,gao2020adversarialnas}, among others. However, even though NAS methods perform well on designing Convolutional Neural Networks (CNNS), they still encounter many problems: 1) the time taken for completing the whole process is in the order of days, and in some cases, months of GPU computing; 2) search spaces are designed by humans, which are usually small and contain complex operations that are comprised of multiple atomic operations, ultimately introducing bias and helping the generated architectures to achieve better results; 3) the size, and regularly, the final architecture of the generated networks is specified by the users; and 4) the inference time of the generated methods is usually poor, due to the fact that generated methods grow both in breadth and depth. 

To mitigate the problems mentioned above, in this work, we propose a NAS method that performs a macro-architecture search, without explicitly defining any outer-skeleton or initial architecture. Moreover, the proposed method can autonomously generate a search space, and at the same time, leverage information about human-designed networks, which were the result of years of expertize, practice and many trial-and-error experiments. To generate the search space, we use all the CNNs implemented in PyTorch, which tremendously increases the operations pool (search space) when compared to other NAS methods. Then, we generate a Hidden Markov Chain for each model and assign a Fitness to it, by evaluating the method using the validation set. 

The search strategy uses Bayesian Optimization to generate networks, which are partially trained and then evaluated in the validation set, which sets their Fitness score.Evolutionary Strategies are also used to allow the system to evolve for several generations, based on the architectures generated in the previous generation. In the end, the best model (highest ranking fitness) is trained from scratch until convergence. The proposed method is capable of generating competitive networks in only a few GPU hours.

The main contributions of this work are:
\begin{itemize}
    \item A novel NAS method, that designs complete architectures from scratch, without requiring human-defined parameters, restricting the number of layers, or specifically forcing an architecture shape or structure;
    \item A simple, yet effective Search Strategy coupled with Bayesian Optimization, that with only a few GPU hours can create, from scratch, competitive CNNs;
    \item A novel way of autonomously designing search spaces, by leveraging information about prior networks to create Hidden Markov Chains that contain information about layer transition and layer components.
\end{itemize}

The remainder of this work is structured as follows: Section \ref{sec:relatedwork}, presents a background an related work study. Section \ref{sec:proposedmethod} presents, in detail, the proposed method. Section \ref{sec:experiments} explains the experiments conducted and the setup. Section \ref{sec:concs} provides a conclusion.

\section{Background and Related Work}
\label{sec:relatedwork}
NAS is a field of research that focuses on developing algorithms for automating the design of Neural Networks, which are widely used in the field of ML. NAS methods are composed of several components: i) the search space, which defines the pool of possible operations and ultimately, the type(s) of networks that can be designed; ii) the search strategy, that defines the approach that is used to explore the search space and generate architectures; and iii) the performance estimation strategy, which is responsible for evaluating the performance of the generated architectures. The architecture search can be done by performing either a micro or a macro-search. In micro-search, methods focus on creating cells or blocks that are replicated multiple times to comply with an outer-skeleton architecture, defined by humans, whereas, in macro-search, NAS methods try to evolve entire network architectures. 

NAS was initially formulated as a Reinforcement Learning (RL) problem, where a controller was trained over-time to sample more efficient architectures \citep{DBLP:journals/corr/ZophL16}. Although this was a novel and significant contribution, it required more than 60 years of computation to solve a particular task. In \citep{Zoph_2018}, a cell-based search in a search space of 13 operations was performed to find reduction cells (reduce input size), and normal cells (perform operations). These cells were then stacked to form entire networks. The authors also found that the final architecture designed using CIFAR-10 \citep{cifar10}, could be successfully transferred to ImageNet \citep{deng2009imagenet} by stacking more cells, achieving state-of-the-art results in both datasets. The downsize was that this method took over 2000 days of computation to achieve good results.

The use of RL as a mechanism for performing NAS is a common approach. In \citep{DBLP:conf/iclr/BakerGNR17}, the authors use Q-learning to train the sampler agent. Using a similar approach, \citep{DBLP:conf/cvpr/ZhongYWSL18} performs NAS by sampling blocks of operations instead of cells, which can then be replicated to form networks. More recently, ENAS \citep{ENAS}, using a controller to discover architectures by searching for an optimal subgraph within a large computational graph, showed that it was possible to use RL, requiring only a few computational days. DARTS, a gradient-based method, showed that by performing a continuous relaxation of the parameters, they could be optimized using a bi-level gradient optimization in a few GPU days \citep{liu2018darts}. This work was then improved using regularization mechanisms \citep{zela2020understanding} and served as the basis for many others \citep{xie2018snas,chen2019progressive, cai2018proxylessnas,Xu2020PC-DARTS:}. Evolutionary-based NAS takes inspiration from biologic systems and is a promising NAS line of research, as it includes methods for evolving state-of-the-art architectures \citep{real2019regularized}, to hierarchically represent architectures and allow easier search \citep{liu2018hierarchical}. 

More in-line with our work, are methods that perform macro-search, such as ENAS \citep{ENAS}, which is an efficient approach to macro-search NAS. However, ENAS contains some drawbacks. First, the number of layers for the final architecture needs to be specified. Finally, the search space is composed of 6 operations: $\{3*3, 5*5\}$ convolutions, $\{3*3, 5*5\}$ depthwise-separable convolutions and max pooling and average pooling with a kernel size of $3*3$, at the same time forcing residual blocks, by sampling 2 previous layers to serve as input to the next one, not allowing for true search and introducing bias. On the other hand, NAS \citep{DBLP:journals/corr/ZophL16}, while capable of generating architectures from scratch, uses a controller that is explicitly parameterized to sample architectures with no less than 6 layers, and to increase that size by 2 every 1600 samples, while taking 22400 GPU days to sample an efficient architecture for CIFAR-10. LEMONADE \citep{elsken2018efficient}, designs architectures by performing three network morphism operations - insert convolution, insert skip connection, and increase the number of filters, requiring an initial architecture, where these operations are then performed.

In this paper, we propose \textbf{HMCNAS}, a method capable of generating a CNN from scratch, in a complex, arbitrarily large, search space, that is automatically created without human-intervention or compounded operations (complex operations that are the result of combining two or more operations), and that does not require any parameterization regarding the number number of layers nor its architecture.


\section{Proposed Method}
\label{sec:proposedmethod}
The proposed method is composed of several parts: i) a novel search space mechanism, which is capable of autonomously creating new search spaces, and in our case, uses information about human-designed networks, since those networks are the result of years of expertise, trial and error, and are known to perform well; and ii) a novel NAS mechanism, that performs Bayesian optimization with evolutionary strategies. In the next sections, we detail the novel approaches proposed in this work: the novel strategy to create search spaces and the new way of representing networks to allow the generation of new architectures using Bayesian optimization.

\subsection{Search Space}
In this work, the search space is composed of the operations contained in human-generated networks that are known to do well. From these, we extract information regarding the layer transition probabilities and create a Hidden Markov chain for each network, where vertices represent Layers and edges represent the probability of going from one layer to the next. A node also contains probabilities for different layer components. The Hidden Markov chains represent the initial population (the search space), forcing the search space to be the possible operations and components that were used in those networks. This means that, in contrast to the most common search spaces that use just a few operations, our method has a search space which is inherited from other methods - requiring no definition or human intervention. Moreover, it does not require the specification of the number of layers that the networks should have, and is extremely flexible, as new networks with different operations can be added with ease.

\begin{figure}[!t]
\centering
\begin{subfigure}[b]{.24\textwidth}
  \centering
  \includegraphics[width=.95\linewidth]{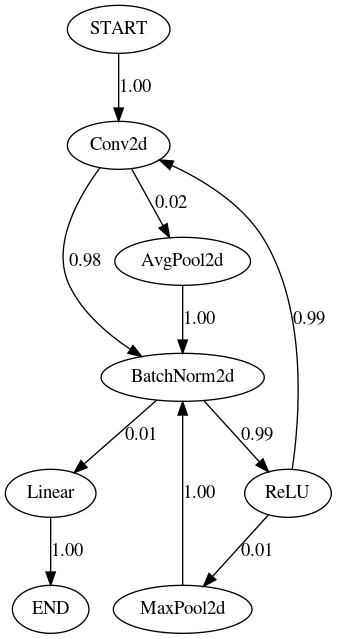}
  \caption{DenseNet-161}
  \label{fig:sfig2}
\end{subfigure}
\begin{subfigure}[b]{.24\textwidth}
  \centering
  \includegraphics[width=.95\linewidth]{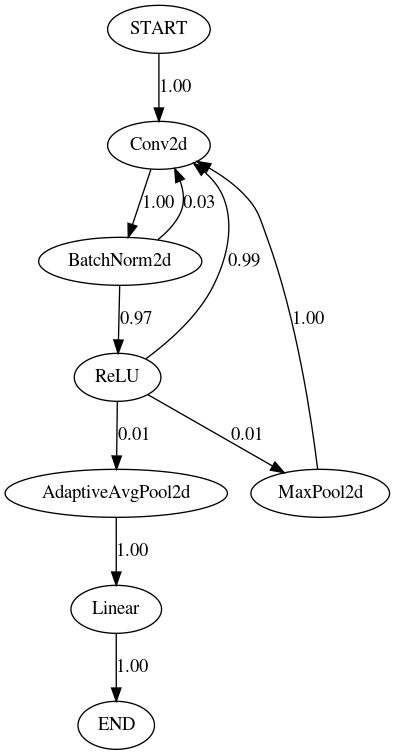}
  \caption{ResNet-152}
  \label{fig:sfig3}
\end{subfigure}
\begin{subfigure}[b]{.24\textwidth}
  \centering
  \includegraphics[width=.95\linewidth]{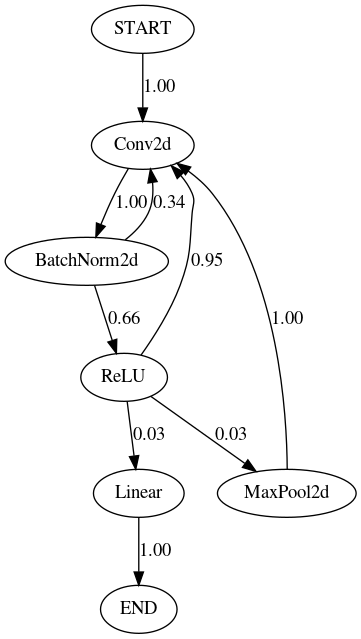}
  \caption{ShuffleNet\_v2}
  \label{fig:sfig4}
\end{subfigure}
\begin{subfigure}[b]{.24\textwidth}
  \centering
  \includegraphics[width=.95\linewidth]{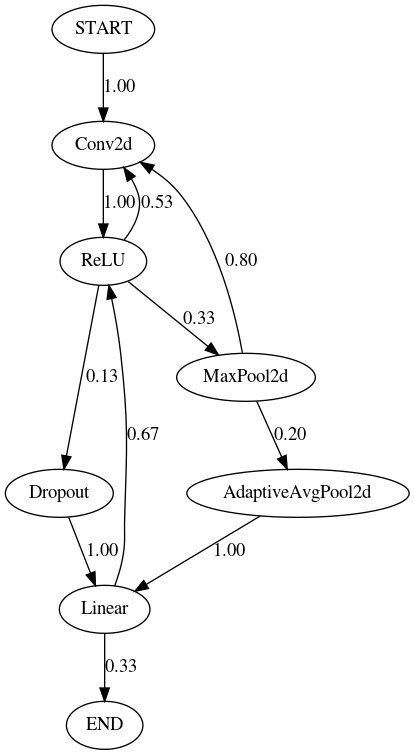}
  \caption{VGG16}
  \label{fig:sfig5}
\end{subfigure}
\caption{Example of generated Hidden Markov-Chains for human-designed models.}
\label{fig:figPytorchMarkov}
\end{figure}

To create the HMCNAS the initial population, we use the implemented CNN models present on TorchVision 1.6: AlexNet \citep{krizhevsky2014one}, GoogLeNet \citep{szegedy2015going}, Inception v3 \citep{szegedy2016rethinking}, VGG \{11, 16, 19\} and their batch normalization versions \citep{Simonyan15}, ResNet \{18, 34, 50, 101, 252\} \citep{he2016deep}, ResNext \{50, 101\} \citep{xie2017aggregated}, DenseNet \{121, 161, 169, 201\} \citep{huang2017densely}, MNASNet \{0\_5, 0\_75, 1\_0, 1\_3\} \citep{tan2019mnasnet}, MobileNet v2 \citep{sandler2018mobilenetv2}, ShuffleNet v2 \{0\_5, 1\_0, 1\_5, 2\_0\} \citep{ma2018shufflenet}, SqueezeNet \{1\_0, 1\_1\} \citep{iandola2016squeezenet}, and Wide ResNet \{50, 101\} \citep{Zagoruyko_2016}, thus totaling 34 models, which results in a search space of 19 different operations. Examples of generated Markov-Chains of PyTorch models are shown in Figure \ref{fig:figPytorchMarkov}. 

Each of the models that comprise the search space were trained for 1 epoch and then had their performance evaluated using the validation set, from which the validation accuracy was extracted to be used as fitness in the upcoming processes. For our experiments, we did this using the CIFAR-10 dataset, in two different setups: using the entire dataset, and a partial dataset, that comprises a stratified 10\% of the entire dataset. The fitness of each model and the time taken to create both search spaces can be seen in Table \ref{tab:seach-space}.


\begin{table}[!tbp]
\centering
\caption{Accuracy of each method that composes the search space, both using entire and partial CIFAR-10 dataset. The last row represents the time, in GPU days, taken to create the entire search space using a GeForce 1080Ti graphics card.}
\label{tab:seach-space}
\begin{tabular}{cc}%
\begin{tabular}[t]{lcc}
\toprule
 \multicolumn{1}{c}{\multirow{2}{*}{\begin{tabular}[c]{@{}c@{}c@{}}\textbf{Model}\end{tabular}}}                                           & \multicolumn{2}{c}{\begin{tabular}[c]{@{}c@{}}\textbf{CIFAR-10} \\ \textbf{Accuracy} (\%)\end{tabular}} \\
 \cmidrule(lr){2-3}
                   & Partial  & Entire \\ \midrule
AlexNet               & 49.60 & 82.92 \\
DenseNet121           & 81.10 & 93.51 \\
DenseNet161           & 87.50 & 95.22 \\
DenseNet169           & 86.20 & 94.15 \\
DenseNet201           & 85.30 & 95.35 \\
GoogLeNet             & 82.70 & 92.94 \\
MNASNet0\_5           & 3.20  & 12.38 \\
MNASNet0\_75          & 9.80  & 16.19 \\
MNASNet1\_0           & 3.20  & 80.89 \\
MNASNet1\_3           & 9.20  & 9.56  \\
MobileNet\_v2         & 79.30 & 90.98 \\
ResNet18              & 82.00 & 92.77 \\
ResNet34              & 79.80 & 93.90 \\
ResNet50              & 87.20 & 94.18 \\
ResNet101             & 89.50 & 95.13 \\
ResNet152             & 88.30 & 95.77 \\
ResNeXt50\_32x4d      & 88.10 & 95.15 \\
ResNeXt101\_32x8d     & 92.20 & 96.56 \\
\end{tabular}
\begin{tabular}[t]{lcc}
\toprule
 \multicolumn{1}{c}{\multirow{2}{*}{\begin{tabular}[c]{@{}c@{}c@{}}\textbf{Model}\end{tabular}}}                                           & \multicolumn{2}{c}{\begin{tabular}[c]{@{}c@{}}\textbf{CIFAR-10} \\ \textbf{Accuracy} (\%)\end{tabular}} \\
 \cmidrule(lr){2-3}
                    & Partial  & Entire \\ \midrule
ShuffleNet\_v2\_x0\_5 & 68.60 & 86.21 \\
ShuffleNet\_v2\_x1\_0 & 78.80 & 91.00 \\
ShuffleNet\_v2\_x1\_5 & 23.30 & 46.53 \\
ShuffleNet\_v2\_x2\_0 & 24.50 & 48.07 \\
SqueezeNet1\_0        & 25.10 & 75.34 \\
SqueezeNet1\_1        & 29.90 & 74.49 \\
VGG11                 & 64.20 & 87.93 \\
VGG11\_bn             & 73.40 & 92.05 \\
VGG13                 & 55.60 & 87.22 \\
VGG13\_bn             & 71.90 & 90.96 \\
VGG16                 & 39.40 & 97.13 \\
VGG16\_bn             & 75.60 & 93.18 \\
VGG19                 & 33.00 & 87.25 \\
VGG19\_bn             & 78.70 & 93.34 \\
Wide ResNet50\_2      & 86.60 & 94.62 \\
Wide ResNet101\_2     & 86.50 & 94.67 \\ \midrule
\multicolumn{1}{r}{$\Delta t$} (days) & 0.016                                     & 0.143 \\ \bottomrule   
\end{tabular}
\end{tabular}

\end{table}

\subsection{Designing Convolutional Neural Networks}
After the initial population has been created, as opposed to most common strategies, which are based on reinforcement learning, gradient descent or evolutionary strategies, the proposed method, uses Bayesian Optimization and an Evolutionary Algorithm to perform the search. In this, a new generation is created by: 1) performing Elitism, where the 15\% best individuals from the previous generation are copied without modifications, and 2) generating new individuals using Bayesian Optimization, where layers are sampled from the individuals of the previous generation based on their fitness. A model is sampled using the roulette wheel algorithm, where models that do not contain the current layer are removed from the pool of selections. After selecting a model, the next layer and the inner components are sampled from all the possible transitions, taken into account the probability of state transition from the parent model. Then, partial training is performed, and the validation accuracy is obtained, which serves as Fitness score. In the end, State Transition Hidden Markov-Chains are generated for all the new individuals and the generation is incremented by one. 

As opposed to other methods that force residual blocks by having every layer input coming from two other layers, we included residual blocks as a design choice. When a convolutional layer is sampled from the search space, there is a small percentage of it being replaced with a residual block, similar to the Bottleneck in ResNet \citep{Zagoruyko_2016}.

The last component of the proposed method is the model selection, where the best individual from the last generation is selected and trained from scratch for several epochs and evaluated using the test set.

\section{Experiments}
\label{sec:experiments}

\subsection{Experimental Setup}
To evaluate the proposed method, we used the following parameters: the number of generations, $n$ was 50, the number of individuals $nI$ per generation was 25. Elitism was set to 15\%, meaning that the 15\% individuals with highest fitness from generation $i$ is passed to generation $i+1$ without any changes or further training in the next generation. The number of epochs to perform the partial train while searching for architectures was 1, and the number of epochs to train the best model after completing all the generations was 100, using Stochastic Gradient Descent \citep{bottou2018optimization}.

To conduct the experiments, we used a computer with an NVidia GeForce GTX 1080 Ti, 16Gb of ram, SSD disk and an AMD Ryzen 7 2700 processor.

\subsection{Results}
In this experiment, we search for entire neural network architectures using the partial CIFAR-10 dataset, where training set was split up in training (4k) and a validation set (1k). In the process of training the final model, the CIFAR-10 dataset was used entirely: 40k images for train, 10k for validation, and the test set remained unchanged, with 10k images. To perform the search, the proposed method took 0.18 days, and the final model, obtained a test error of 9.41\%. The results are compared against the state-of-the-art in Table \ref{tab:my-tableresults}, were the first block represents a high performant human-designed network, and the second block presents the results of different approaches that design entire networks. Note that, while HMCNAS has a higher error rate, it is extremely fast, requiring only 0.23 days to search for a competitive model. Moreover, it was done using a more complex search space, without any required human-defined parameters, as opposed to the other methods that require a definition of initial architectures, the number of layers to be sampled, and rely on small and biased search spaces. Thus, HMCNAS is a promising approach to generalize NAS, as for many problems (including new and unseen ones), it may be the only viable option, as the need for defined parameters relies on human knowledge about the specific problem.

\begin{table}[!t]
\renewcommand{\arraystretch}{1.25}
\centering
\caption{Classification errors of different methods on CIFAR-10. The first block presents state-of-the-art human-designed networks. The second block presents approaches that design the entire networks. Acronyms used: RL - Reinforcement Learning, OS - One Shot, EA - Evolutionary Algorithm, GB - Gradient Boost, BO - Bayesian Optimization, R - Random.}
\label{tab:my-tableresults}
\resizebox{\columnwidth}{!}{%
\begin{tabular}{lcccc}
\toprule
\textbf{Architecture} & \begin{tabular}[c]{@{}c@{}}\textbf{Test Error}\\ (\%)\end{tabular} & \begin{tabular}[c]{@{}c@{}}\textbf{Search Cost}\\ (GPU Days)\end{tabular} & \begin{tabular}[c]{@{}c@{}}\textbf{Params}\\ (M)\end{tabular} & \begin{tabular}[c]{@{}c@{}}\textbf{Search}\\ \textbf{Method}\end{tabular} \\
\midrule
DenseNet-BC \citep{huang2017densely} & 3.46 & - & 25.6 & manual \\
\midrule
\midrule
Budgeted Super Nets \citep{veniat2018learning} & 9.21 & - & - & - \\
ConvFabrics \citep{saxena2016convolutional} & 7.43 & - & 21.2 & - \\
Macro NAS + Q-Learning \citep{baker2016designing} &  6.92 & 80-100 & 11.2 & RL \\
Net Transformation \citep{cai2018efficient} & 5.7 & 10 & 19.7 &  RL \\
SMASH \citep{brock2017smash} & 4.03 & 1.5 & 16.0 & OS \\
NAS \citep{DBLP:journals/corr/ZophL16} & 4.47 & 22400 & 7.1 & RL \\
NAS + more filters \citep{DBLP:journals/corr/ZophL16} & 3.65 & 22400 & 37.4 & RL \\
ENAS \citep{ENAS} & 3.87 & 0.32 & 38.0 & RL \\
LEMONADE I \citep{elsken2018efficient} & 3.37 & 56 & 8.9 & EA \\
RandGrow \citep{humacro} & 2.93 & 6 & 3.1 & R \\
Petridish \citep{hu2019efficient} & 2.83 & 5 & 2.2 & GB \\

\textbf{HMCNAS} & 9.41 & 0.23 & 5.1 & EA+BO
\\ \bottomrule
\end{tabular}
}
\end{table}

\section{Conclusions}
\label{sec:concs}
In this work, we propose a novel approach to the problem of neural architecture search for designing entire architectures from scratch. Our method is capable of automatically designing search spaces, and leverages human knowledge by using information about some of the best Human-designed networks, which were the result of years of experience and trial and error. By using such information, we can create state transition Hidden Markov-chains of each one of the networks and use that to generate new architectures.

By using a Bayesian selection, the method is extremely fast and even though it gives more importance to the best networks of the previous generation, it allows for controlled novelty search by also selecting components from networks with smaller fitnesses. The proposed method not only removes the need to specify the possible operations, but also the number of nodes (layers) to be sampled, and information about initial architectures or final skeletons. HMCNAS provides a step towards generalizing NAS, by providing a way to create competitive models, without requiring any human knowledge about the specific task.


\bibliography{bibliography}
\bibliographystyle{iclr}


\end{document}

%% file: math_commands.tex
\usepackage{amsmath,amsfonts,bm}

\def\eqref#1{equation~\ref{#1}}

\def\1{\bm{1}}

\DeclareMathAlphabet{\mathsfit}{\encodingdefault}{\sfdefault}{m}{sl}
\SetMathAlphabet{\mathsfit}{bold}{\encodingdefault}{\sfdefault}{bx}{n}

